\documentclass{article}
\newif\ifanonymous \anonymousfalse   
\usepackage{xcolor}
\ifanonymous\usepackage{corl_2026}\else\usepackage[final]{corl_2026}\fi
\makeatletter
\renewcommand{\@noticestring}{Accepted at the CoRL 2026 Workshop ``Do Robots Need World Models?'', Austin, TX, USA.}
\makeatother
\usepackage{amsmath,amssymb,amsthm}
\usepackage{graphicx}
\usepackage{booktabs}
\usepackage{array}
\newcolumntype{L}[1]{>{\raggedright\arraybackslash}p{#1}}
\usepackage{microtype}
\hypersetup{pdfauthor={Hassan Iqbal and Krishna Kumar},pdftitle={What Must a Model Tell a Planner? Outcome Interfaces for Risk-Aware Planning After Robot Damage}}

\usepackage{tikz}
\theoremstyle{remark}

\definecolor{dec}{HTML}{C55A11}
\definecolor{rto}{HTML}{2E7D32}
\definecolor{con}{HTML}{1F4E79}


\title{What Must a Model Tell a Planner?\\Outcome Interfaces for Risk-Aware Planning After Robot Damage}
\author{
Hassan Iqbal\\
The University of Texas at Austin\\
\texttt{hassan.iqbal@utexas.edu}
\And
Krishna Kumar\\
The University of Texas at Austin\\
\texttt{krishnak@utexas.edu}
}
\begin{document}
\maketitle

\begin{abstract}
When a robot is damaged, it must decide which routes it can still afford, and the amount of risk it may accept can change from one mission to the next.
We argue that whether a learned model helps such a robot depends on its decision interface, that is, on what the model hands to the planner.
A decision policy tells the planner what to do, whereas an outcome model, a kind of world model, predicts the consequences of each route.
We study simulated Unitree Go2 quadrupeds and H1 humanoids with weakened motors, added payloads, or reduced foot friction.
We compare four interfaces.
The first is a decision policy, and the second is the same policy trained with fixed penalties for infeasible routes.
The third is an outcome model that predicts each route's feasibility and travel time.
The fourth, which we call a capability contract, is an outcome model that predicts the robot's chance of success and its travel time on each terrain and composes them along a route.
All share the same collected experience, matched tuning and risk calibration, and an uncertainty guard.
Two comparatively low-risk probes let the capability contract rank hazardous terrains about as well as four new trials on every hazard.
At a calibrated 5\% limit on selecting empirically infeasible routes, outcome models cut declined missions from about 32\% to 7--8\% on Go2 and from 57\% to 35--36\% on H1 compared with the plain policy.
Fixed penalties narrow this gap, but the best penalty changes with the risk target.
Predicting outcomes lets the planner preserve missions by changing routes as risk requirements tighten.
\end{abstract}


\section{Introduction}
Consider a quadruped whose front-left knee has lost half of its torque.
Its planner can send it up a short flight of stairs, along a mixed path with a single obstacle, or around a long detour over gentle slopes, and the robot can always stop and ask for help.
The terrain and available routes have not changed, but the robot's chances of completing them have.
Testing the hazards by trial would invite the falls the planner exists to prevent.
The acceptable risk may change from one mission to another, with stricter requirements near people than during a survey.
If the preferred route becomes unacceptable, the planner must still decide which alternative the damaged robot can attempt.

Policies map observations to choices~\citep{chi2025diffusion,zitkovich2023rt,kim2024openvla}, while world models predict consequences that can inform those choices~\citep{ha2018recurrent,hafner2023mastering}.
We study their \textit{decision interface}, what the learned component hands to the planner.
A \textit{decision policy} scores routes, while an \textit{outcome model} estimates feasibility or success probability and travel time.
Confidence that a route is preferred is not the same as the probability that the robot can complete it.

\begin{figure}[t]
\centering
\resizebox{0.92\textwidth}{!}{%
\begin{tikzpicture}[font=\footnotesize, x=1cm, y=1cm]
  \definecolor{rd}{HTML}{8B1E3F}\definecolor{rm}{HTML}{B8860B}\definecolor{re}{HTML}{6B8E23}
  \coordinate (S) at (0,0); \coordinate (G) at (3.4,0);
  \draw[line width=1.6pt, color=rd] (S) .. controls (0.85,1.3) and (2.55,1.3) .. (G);
  \draw[line width=1.6pt, color=rm] (S) -- (G);
  \draw[line width=1.6pt, color=re] (S) .. controls (0.55,-1.6) and (2.85,-1.6) .. (G);
  \fill (S) circle (2.2pt); \fill (G) circle (2.2pt);
  \node[left] at (S) {start}; \node[right] at (G) {goal};
  \node[text=rd] at (1.7,1.3) {direct -- stairs, boxes, rough};
  \node[text=rm] at (1.7,0.22) {mixed -- boxes, slopes};
  \node[text=re] at (1.7,-1.42) {detour -- six slopes};
  \tikzset{box/.style={draw, rounded corners=2pt, line width=0.9pt, align=left, text width=8.75cm, inner sep=3.5pt, anchor=west}}
  \node[box, draw=dec] at (4.75,1.55) {\textbf{\textcolor{dec}{Decision policy}} tells the planner which route an oracle would take\\
    mixed 0.62 \quad detour 0.25 \quad direct 0.03 \quad fall back 0.10\\
    \emph{Stricter target} demand more confidence, so the robot falls back.};
  \node[box, draw=rto] at (4.75,0.0) {\textbf{\textcolor{rto}{Route-outcome model}} tells the planner each route's feasibility score\\
    direct 0.41 \quad mixed 0.84 \quad detour 0.97\\
    \emph{Stricter target} drop the mixed route and take the detour.};
  \node[box, draw=con] at (4.75,-1.55) {\textbf{\textcolor{con}{Capability contract}} tells the planner each terrain's chance of success\\
    stairs 0.71 \quad boxes 0.85 \quad rough 0.68 \quad slope 0.995\\
    \emph{Stricter target} take the detour using composed estimates.};
\end{tikzpicture}}
\caption{Illustrative predictions for one mission.
The decision rule accepts its preferred route or falls back, whereas outcome planners can select a qualifying route.
The route model predicts route feasibility, while the contract predicts terrain success and composes estimates for new combinations.}
\label{fig:concept}
\end{figure}
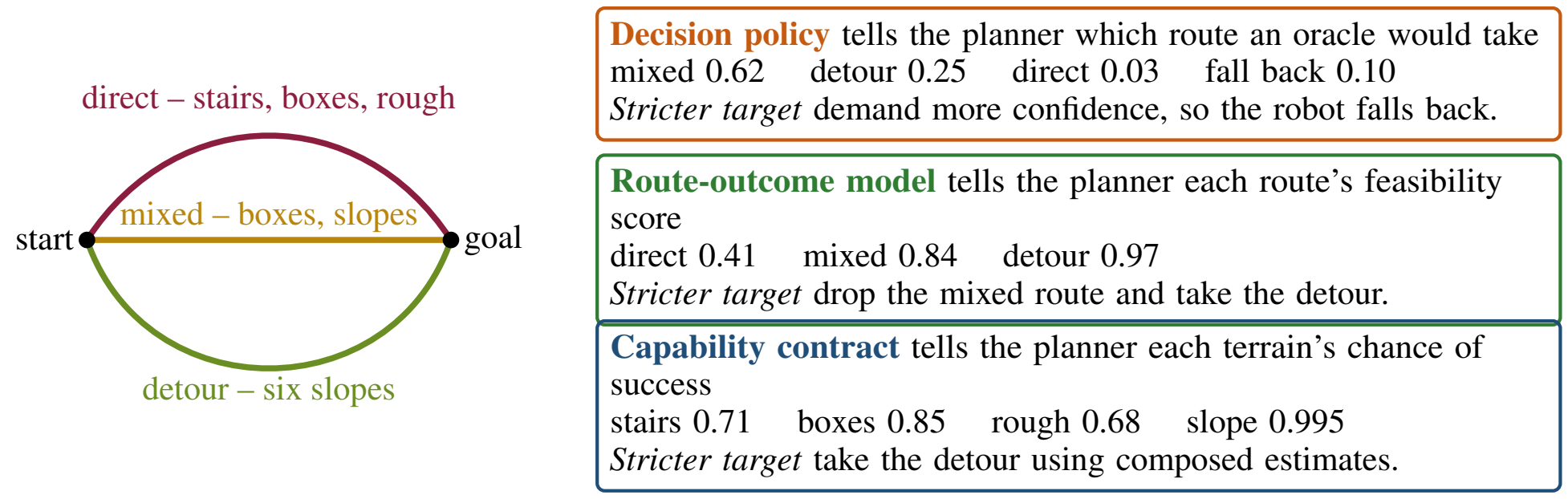

Tightening the risk target reveals a practical difference between the evaluated rules (Fig.~\ref{fig:concept}).
The decision policy accepts its preferred route or falls back, whereas the outcome planners can select another qualifying route.
We compare four interfaces on damaged Go2 and H1 robots with matched learning and calibration.

\paragraph{Related work.}
Damage recovery uses Gaussian process predictions to search for compensatory behaviors~\citep{cully2015animals}, while adaptive locomotion learns how to act under changed dynamics~\citep{kumar2021rma,liu2025locoformer}.
Whereas these methods improve how the robot moves, we assess where its existing controller can still succeed.
Related assessment methods use competence estimates to decide when autonomy should defer to human assistance~\citep{basich2023competence}, or predict perception performance from experience to avoid failure~\citep{gurau2018experience}.
Our Gaussian process instead learns across damaged robots how responses on easier terrain predict the probability of successful traversal and travel time on hazards.
This relationship lets two comparatively low-risk probes assess a new robot without retesting every hazard.
We evaluate the resulting hazard ranking and the trials and falls required to obtain it against direct hazard retesting.

Model-free RL can learn locomotion without an explicit transition model~\citep{rudin2022minutes}, while model-based RL can combine policy proposals with predicted rollouts~\citep{hansen2024td,zhao2025moe}.
These approaches establish that learned policies and outcome predictions can play complementary roles.
We ask what a route planner gains when information about physical execution remains explicit rather than being compressed into a route choice.
Navigation methods already incorporate deployment instructions~\citep{potnis2026catnav,weerakoon2025behav}, and conformal planners calibrate uncertain predictions or decisions~\citep{lindemann2023safe,dixit2023adaptive,lekeufack2024cdt,ren2023knowno}.
With shared locomotion, experience, and risk calibration, we compare filtering a learned route choice with reevaluating alternatives from route or terrain predictions.
This comparison measures the missions retained as risk targets change and tests whether predictions for individual terrains help when missions recombine them.

\section{Setting and Interfaces}
\label{sec:setting}
A \emph{context} $x$ specifies a terrain type, difficulty, and observable start state.
For robot $r$, $p_r(x)$ is the empirical success fraction over repeated traversals and $d_r(x)$ is the measured duration given success.
A route $\rho$ is a sequence of such traversals $e$.
The benchmark composes their outcomes as $P_r(\rho)=\prod_{e\in\rho}p_r(e)$ and $D_r(\rho)=\sum_{e\in\rho}d_r(e)$.
A mission offers three routes and a fallback that declines traversal.
We label a route feasible when $P_r(\rho)\ge p_{\min}=0.7$.
The loss $L_r$ is the fraction of all missions on which an empirically infeasible route is selected, with fallback contributing zero.
The risk target $\alpha$ bounds the expectation of this loss under the calibration assumptions.
A 5\% target therefore limits empirically infeasible selections to at most 5\% of all offered missions in expectation, including declined missions in the denominator.
It does not bound falls, mission failure, or the error fraction conditional on committing to a route.
Declining every mission has zero loss, so the useful comparison is how many missions each interface retains at the same calibrated target.

After damage, the robot performs two comparatively low-risk probes.
Their observation vector $z$ records success, duration, energy, changes in speed and pitch, and per-leg torque saturation.
All four interfaces use these observations and the same collected traversal experience, with kernel regression across robots and matched tuning.
The \emph{decision policy} learns scores for the oracle's choice among three routes and fallback.
The oracle selects the fastest empirically feasible route using measured traversal outcomes, not Gaussian-process predictions.
The \emph{penalized decision policy} subtracts a fixed penalty $P$ from training scores for infeasible routes.
The \emph{route-outcome model} learns each route's feasibility label and duration.
The \emph{capability contract} predicts terrain success and duration, with a Gaussian process~\citep{williams2006gaussian}, and composes these estimates along routes.
The contract learns how probe observations predict hazardous-terrain outcomes from paired measurements on fitting robots.
Probes are chosen greedily to reduce error in predicted success.
Route and mission descriptions, supervision, and tuning appear in Appendix~\ref{app:e2e}.

Each planner uses an adjustable decision margin.
The contract compares composed success estimates with $p_{\min}$ plus the margin, while the route model compares its feasibility scores with $0.5$ plus the margin.
Each takes the fastest admitted route, or falls back if none qualifies.
The decision planner commits only when its top route score exceeds the fallback score by the margin, so increasing the margin can only replace commitments with fallback.
Conformal risk control calibrates each margin on 100 robots~\citep{angelopoulos2024conformal}.
Changing $\alpha$ changes this calibration, while $p_{\min}$ remains fixed.
A shared guard uses predictive variance over probe observations to send unfamiliar robots to fallback.
Appendix~\ref{app:guarantee} states the assumptions and the expected-loss guarantee.

\begin{figure}[t]
\centering
\includegraphics[width=\textwidth]{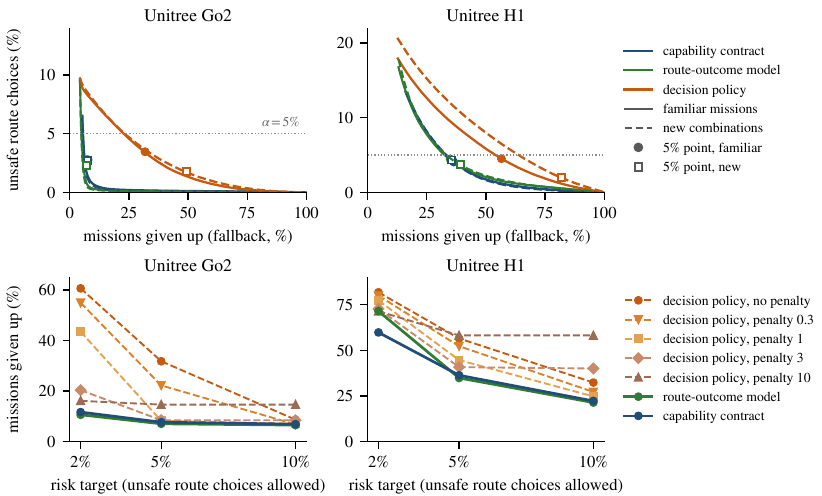}
\caption{Infeasible selections and declined missions.
Above, increasing the margin traces their tradeoff, with markers at a calibrated 5\% selection-risk target.
Below, the best fixed penalty changes with the risk target.
\emph{Unsafe} in the axes means empirically infeasible selection, not observed execution failure.
Curves average 20 splits.}
\label{fig:results}
\end{figure}

\section{Benchmark}
\begin{figure}[!t]
\centering
\begin{minipage}{0.235\textwidth}
\centering
\includegraphics[width=\linewidth,trim=147.456bp 241.920bp 442.368bp 2.880bp,clip]{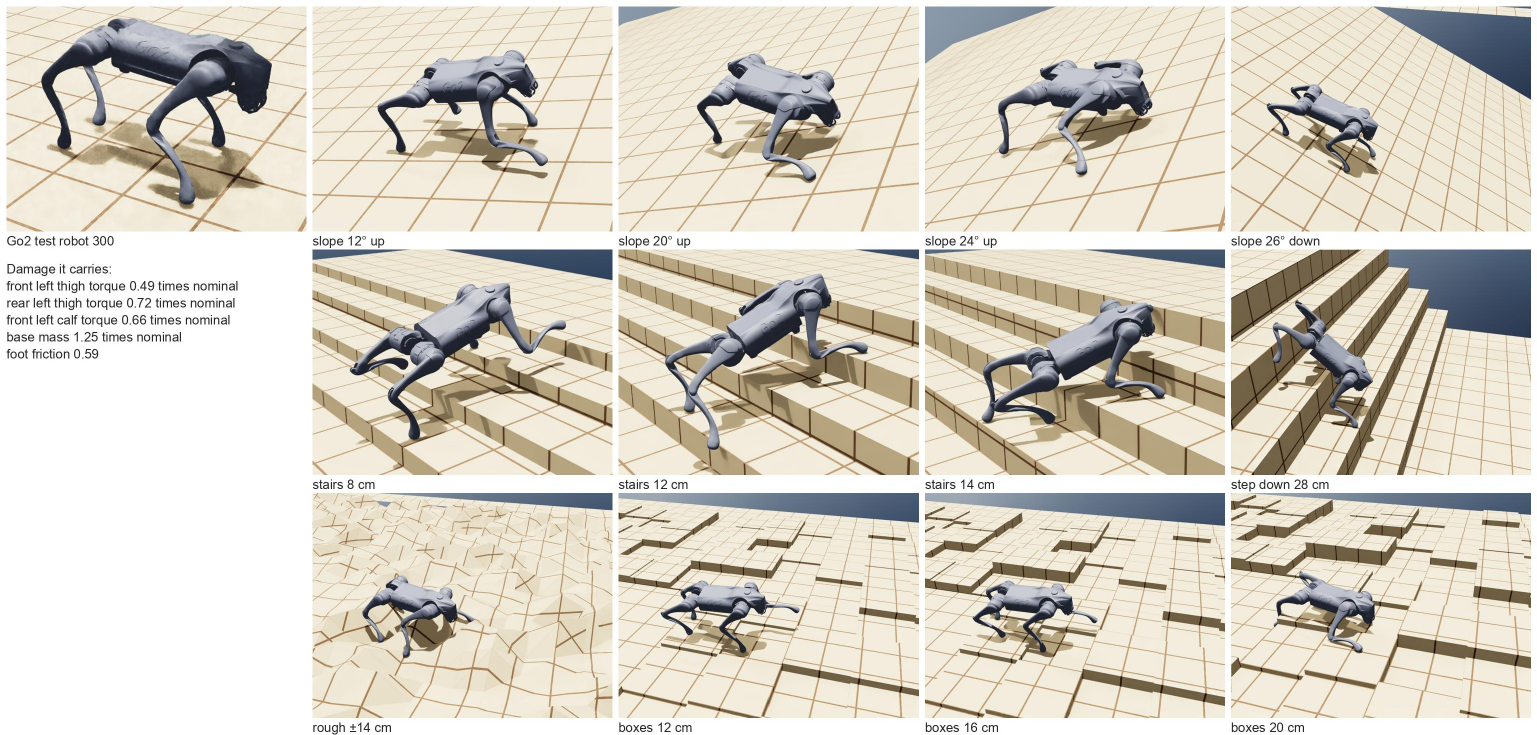}
{\footnotesize Go2, $12^\circ$ ascent\par Probe candidate\par}
\end{minipage}\hfill
\begin{minipage}{0.235\textwidth}
\centering
\includegraphics[width=\linewidth,trim=294.912bp 125.280bp 294.912bp 119.520bp,clip]{figures/go2_terrain_examples.pdf}
{\footnotesize Go2, 12\,cm stairs\par Hazard\par}
\end{minipage}\hfill
\begin{minipage}{0.235\textwidth}
\centering
\includegraphics[width=\linewidth,trim=442.368bp 241.920bp 147.456bp 2.880bp,clip]{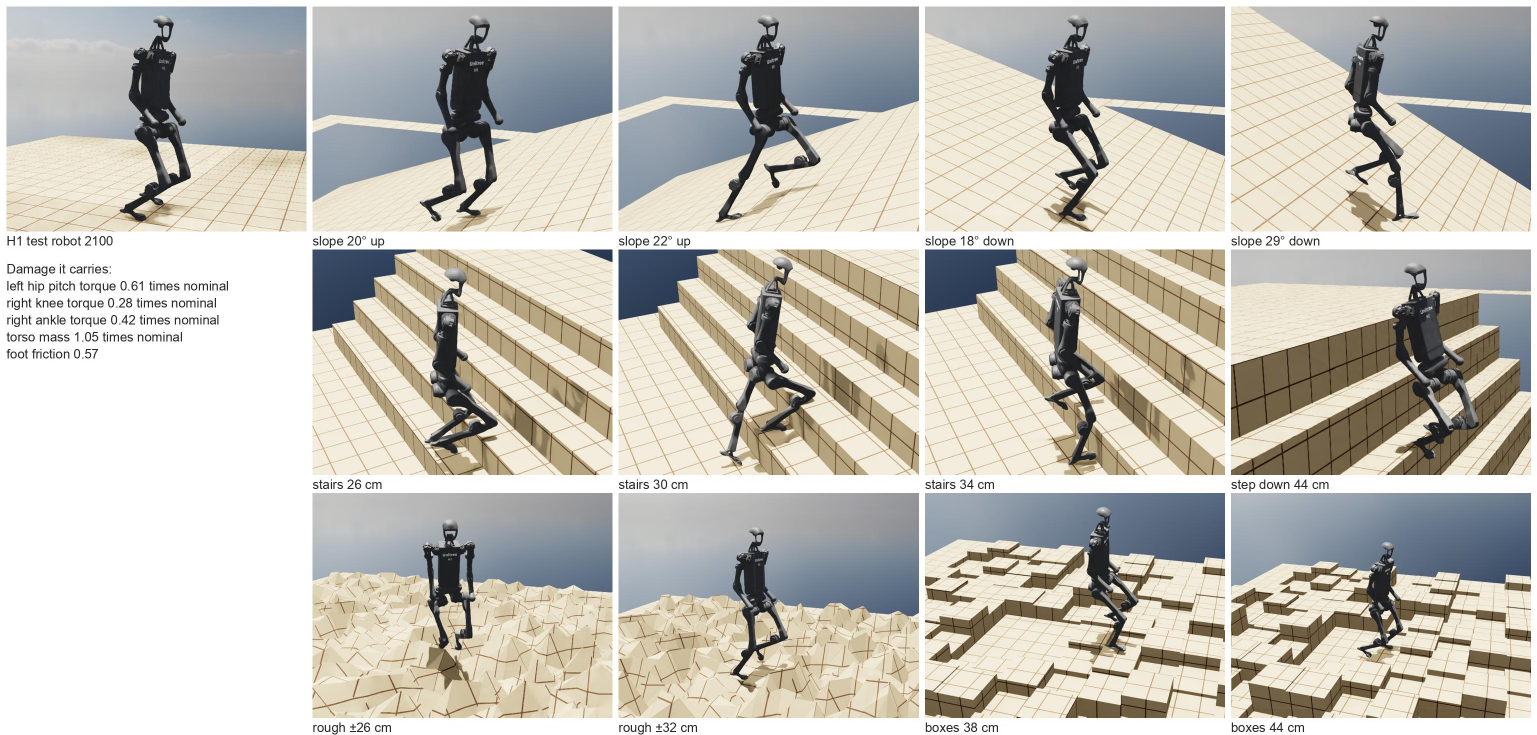}
{\footnotesize H1, $18^\circ$ descent\par Probe candidate\par}
\end{minipage}\hfill
\begin{minipage}{0.235\textwidth}
\centering
\includegraphics[width=\linewidth,trim=294.912bp 125.280bp 294.912bp 119.520bp,clip]{figures/h1_terrain_examples.pdf}
{\footnotesize H1, 30\,cm stairs\par Hazard\par}
\end{minipage}
\caption{Recorded simulation frames on probe candidates and hazardous stairs after motor, mass, and friction perturbations (Appendix~\ref{app:examples}).}
\label{fig:terrain_examples}
\end{figure}

We use Isaac Lab~\citep{mittal2023orbit} with frozen pretrained Go2 and H1 locomotion policies.
Each robot has up to three weakened joints, added payload, and reduced foot friction.
Each platform has 300 source robots, split into 200 fitting and 100 calibration robots, and 100 separate test robots.
Go2 adds 80 robots with damage beyond the source population.
Every terrain is traversed from eight observable start states, each repeated with two hidden joint-posture perturbations.
With two trials per context and $p_{\min}=0.7$, a route is feasible exactly when both trials succeed on every edge.
Missions offer a direct route with three hazardous edges, a mixed route with one hazardous and three easy edges, and a detour with six easy edges.
We evaluate familiar missions, new combinations of represented hazardous terrains, and more severe Go2 damage, averaging results over 20 source splits.
The main text reports the familiar-mission risk comparison and probe assessment, while Appendix~\ref{app:more} reports the remaining evaluations.
These route evaluations compose recorded terrain outcomes rather than execute complete routes continuously.
Figure~\ref{fig:terrain_examples} shows the simulated terrains, and Appendices~\ref{app:bench} and~\ref{app:examples} provide benchmark and damage details.

\section{Results}
\label{sec:results}
\paragraph{Outcome models retain more missions under risk limits.}
Both outcome interfaces retain more missions than the plain policy on Go2 and H1 under calibrated risk limits, despite similar agreement with the oracle when no risk target is imposed (Fig.~\ref{fig:results}, Table~\ref{tab:h2h}).
A well-chosen fixed penalty recovers much of this benefit, nearly closing the gap on Go2 and narrowing it on H1, but the best penalty changes with the risk target (Table~\ref{tab:penalty}).
At the strictest target, the contract still retains more missions than the best tested penalty on both platforms.
The practical benefit is that the planner can change routes as requirements change while reusing the same outcome predictor.

\paragraph{Two probes improve hazard assessment at low trial cost.}
Observing the damaged robot improves the ranking of hazardous terrains over the population average.
Two designed probes raise AUC from 0.67 to 0.87 on Go2 and from 0.75 to 0.86 on H1 (Table~\ref{tab:ident}).
Four trials on every hazard achieve similar ranking but require 16--18 times more runs and produce roughly 16--18 falls per robot, compared with fewer than 0.1 during probing.
Increasing the number of designed probes beyond two yields little further improvement (Table~\ref{tab:probes}).

\section{Discussion and Future Work}
Two comparatively low-risk probes let a damaged robot assess hazardous terrain without retesting every hazard itself.
The outcome interfaces then let the evaluated planners reconsider alternatives and retain more missions than the plain policy under calibrated risk limits.
At deployment, we can change the risk target by recalibrating the planner's margin without refitting its outcome predictor, while the contract can compose predictions for new combinations of represented terrains.

A risk-conditioned policy can instead take the requested target as an input and learn the route switch directly, but the policies evaluated here do not receive that target.
Our comparison therefore establishes the value of exposing reusable outcome estimates to this planner, rather than showing that changing risk requirements intrinsically requires an outcome model.

As risk requirements tighten, the planner becomes increasingly conservative, trading mission retention for fewer infeasible route selections.
Future work should preserve useful route choices at stringent targets, validate complete-route execution on physical robots, and test assessment and calibration under broader shifts in robot damage and mission distributions.

\clearpage
\bibliography{refs}

\appendix
\section{Calibration and Risk Interpretation}
\label{app:guarantee}
For each robot $r$ and decision margin $\lambda$, $L_r(\lambda)\in[0,1]$ is the fraction of missions on which the underlying decision rule selects an empirically infeasible route.
Fallback has zero loss.
Let $\Lambda$ be the finite ordered margin grid, including an endpoint $\lambda_{\max}$ that always selects fallback.
Because an outcome planner may switch routes as the margin grows, its original loss need not decrease monotonically.
We therefore form the upper envelope separately for every robot,
\begin{equation}
\widetilde L_r(\lambda)=\max_{\lambda'\in\Lambda,\ \lambda'\ge\lambda}L_r(\lambda').
\end{equation}
With $n=100$ calibration robots, the calibrated margin is
\begin{equation}
\widehat\lambda=\min\left\{\lambda\in\Lambda:
\frac{\sum_{i=1}^{n}\widetilde L_i(\lambda)+1}{n+1}\le\alpha\right\}.
\label{eq:crc}
\end{equation}
If no margin qualifies, the rule uses $\lambda_{\max}$.
For the tested targets $\alpha\in\{0.02,0.05,0.10\}$, the endpoint has zero loss and makes the set nonempty.

Conditional on the fitting data, we assume that calibration robots and a new robot $r^\star$ are exchangeable, that probe design, feature transformations, model fitting, and tuning are fixed before calibration, and that losses use a common fixed mission distribution or shared mission bank.
The envelopes are bounded by one, non-increasing, and at least as large as the original losses.
Applying conformal risk control gives $\mathbb{E}[L_{r^\star}(\widehat\lambda)]\le\alpha$ under these assumptions~\citep[Corollary~1]{angelopoulos2024conformal}.
Conditional on the fitting data, the expectation is over the calibration robots and the new robot, and over mission sampling when missions are not drawn from a fixed shared bank.
This is an application of the published result, not a new theoretical guarantee.

The uncertainty guard compares the new robot's probe variance $s^2(z_{r^\star})$ with a calibration quantile $\tau$.
For $\delta=0.05$, $\tau$ is the $\lceil(n+1)(1-\delta)\rceil$-th smallest calibration variance.
The guard replaces a route selection with fallback when $s^2(z_{r^\star})>\tau$.
Applied after calibration of the underlying rule, it cannot increase $L_{r^\star}$ and therefore preserves the bound.
Under exchangeability, its marginal flag probability is at most $\delta$~\citep{vovk2005algorithmic}.
This statement does not guarantee detection of robots outside the source distribution.

The controlled loss concerns selection, not the physical probability of falling or failing a mission, and it averages over all missions rather than only accepted ones.
The guarantee is marginal, so the observed test loss for an individual calibration split can exceed $\alpha$.
For example, at the 5\% H1 target, the largest observed split rates were 5.7\% for the contract and 5.8\% for the decision policy.
The guarantee does not extend to the new-combination mission distribution or the robots damaged beyond the source population.

\section{Interfaces and Implementation}
\label{app:e2e}
\paragraph{Capability contract.}
A Gaussian process over fitting robots predicts empirical success and duration for each terrain context~\citep{williams2006gaussian}.
For either target at context $x$, its prediction and shared probe variance are
\begin{equation}
\begin{split}
\mu_x(z)&=\bar y_x+k(z)^\mathsf{T}(K+\sigma^2 I)^{-1}(y_x-\bar y_x\mathbf{1}),\\
s^2(z)&=k(z,z)-k(z)^\mathsf{T}(K+\sigma^2 I)^{-1}k(z).
\end{split}
\label{eq:gp}
\end{equation}
Here $Z$ contains the fitting robots' standardized probe observations, $K=k(Z,Z)$ is their kernel matrix, and $k(z)$ is the vector of kernel evaluations between the new observation $z$ and $Z$.
The vector $y_x$ contains the fitting targets, $\bar y_x$ is their mean, $\mathbf{1}$ is the all-ones vector, $I$ is the identity matrix, and $\sigma^2$ is the diagonal regularization parameter.
The contract composes terrain predictions as $\widehat P(\rho)=\prod_{e\in\rho}\widehat p(e)$ and $\widehat D(\rho)=\sum_{e\in\rho}\widehat d(e)$.
Probe selection greedily reduces prediction error for success on hazardous terrain using source-robot experience.

\paragraph{Descriptions and supervision.}
A route is described by counts of edges in each context and terrain, with $96+12$ entries on Go2 and $128+16$ on H1.
A mission stacks the descriptions of its three routes.
The decision learners receive these descriptions and $z$, so they already have observations of the damaged robot.
Their oracle uses measured terrain outcomes to choose the feasible route of least composed duration, or fallback if no route is feasible.
The plain policy learns oracle-choice scores; each penalized policy subtracts $P\in\{0.3,1,3,10\}$ from the training score of every infeasible route.
Each penalty is trained separately and evaluated at all risk targets.
The route-outcome model learns feasibility and duration directly, tuning them separately.
Its feasibility score is clipped to $[0,1]$ and must exceed $0.5$ plus the calibrated margin; the grid extends to $0.51$, where every route is rejected.

\paragraph{Fitting and tuning.}
Each split uses 200 fitting robots and 100 calibration robots, with 100 separate test robots.
All interfaces use kernel regression across robots, with exact leave-one-robot-out residuals for tuning so a held-out robot contributes none of its items to fitting.
The grid uses six robot-kernel length scales and ridge penalties $\{0.03,0.1,0.3,1\}$.
The route and decision models additionally use linear or squared-exponential item kernels with scales of $0.5$, $1$, or $2$ times the median pairwise item distance.
The fitting data and procedures are matched, while the targets, route representations, and selection rules differ.
The probe variance supplies the common uncertainty guard.

\subsection{Robustness of the Policy Comparison}
\label{app:robust}
The decision policy selects the largest ridge penalty in the original grid, motivating a check with penalties up to 100 and item length scales down to 0.125.
This changes its fallback rates by at most 2.3 percentage points and its infeasible-selection rates by at most 1.0 point.
A neural policy with hidden layers of 256 and 128 units, ReLU activations, Adam optimization, and early stopping uses the same inputs, labels, decision margin, and guard on the first five splits.
Its infeasible-selection rates are at most 4.6\%, and it declines more familiar missions than the kernel policy.
A further check changes $p_{\min}$ to $0.5$, which changes feasibility labels and oracle choices, so every interface is retrained and recalibrated.
Table~\ref{tab:robust} reports these comparisons.
They test the sensitivity of the specified policy family, without substituting for a policy conditioned on the requested $\alpha$.

\begin{table}[htbp]
\caption{Declined missions at $\alpha=5\%$, in percent, with the common guard.
Each pair compares the contract (Con.) with the named variant (Var.) on the same splits.
Fam.\ and New denote familiar missions and new terrain combinations.}
\label{tab:robust}
\centering\footnotesize
\setlength{\tabcolsep}{3.2pt}
\begin{tabular}{@{}lrrrr@{\hspace{10pt}}rrrr@{}}
\toprule
 & \multicolumn{4}{c}{Unitree Go2} & \multicolumn{4}{c}{Unitree H1}\\
\cmidrule(r){2-5}\cmidrule(l){6-9}
 & \multicolumn{2}{c}{Fam.} & \multicolumn{2}{c}{New} & \multicolumn{2}{c}{Fam.} & \multicolumn{2}{c}{New}\\
\cmidrule(r){2-3}\cmidrule(lr){4-5}\cmidrule(lr){6-7}\cmidrule(l){8-9}
Variant & Con. & Var. & Con. & Var. & Con. & Var. & Con. & Var.\\
\midrule
Main comparison (20 splits) & 7.7 & 31.8 & 7.5 & 49.2 & 36.4 & 56.6 & 35.4 & 82.0 \\
Kernel policy, wider grid (20 splits) & 7.7 & 31.8 & 7.5 & 49.1 & 36.4 & 55.5 & 35.4 & 81.3 \\
Neural policy (5 splits) & 7.5 & 46.5 & 7.4 & 55.4 & 36.8 & 66.7 & 36.2 & 70.7 \\
Feasibility threshold $p_{\min}=0.5$ (20 splits) & 6.9 & 45.9 & 6.8 & 64.3 & 16.7 & 56.3 & 17.0 & 84.6 \\
Route-outcome model (20 splits) & 7.7 & 7.1 & 7.5 & 7.1 & 36.4 & 34.9 & 35.4 & 39.3 \\
 
\bottomrule
\end{tabular}
\end{table}
\section{Supplementary Results}
\label{app:more}
\subsection{Route Selection and Fixed Penalties}
Table~\ref{tab:h2h} reports infeasible selections and declined missions for the three principal interfaces.
Table~\ref{tab:penalty} adds the four fixed penalties.
The best tested penalty on Go2 increases from 0.3 to 1 to 10 as the target tightens from 10\% to 5\% to 2\%; on H1 it increases from 1 to 3 to 10.
The same outcome predictor serves every target, with its decision margin recalibrated.
The H1 route model's 71.5\% fallback at the 2\% target includes 6 of 20 splits that reject every route.
The absence of a universal advantage for either outcome representation is part of the comparison.

\begin{table}[htbp]
\caption{Main route-selection comparison on separate test robots, averaged over 20 splits.
Uns.\ is the fraction of all missions with an empirically infeasible selection, and Fb.\ is the fraction declined, both in percent.
Bold marks the fewest declined missions for each target.
Calibration applies to familiar missions and robots from the source population.}
\label{tab:h2h}
\centering\footnotesize
\setlength{\tabcolsep}{2.4pt}
\begin{tabular}{@{}lrrrrrr@{\hspace{9pt}}rrrrrr@{}}
\toprule
 & \multicolumn{6}{c}{Unitree Go2} & \multicolumn{6}{c}{Unitree H1}\\
\cmidrule(r){2-7}\cmidrule(l){8-13}
 & \multicolumn{2}{c}{Contract} & \multicolumn{2}{c}{Route-outc.} & \multicolumn{2}{c}{Decision} & \multicolumn{2}{c}{Contract} & \multicolumn{2}{c}{Route-outc.} & \multicolumn{2}{c}{Decision}\\
\cmidrule(r){2-3}\cmidrule(lr){4-5}\cmidrule(lr){6-7}\cmidrule(lr){8-9}\cmidrule(lr){10-11}\cmidrule(l){12-13}
Condition & Uns. & Fb. & Uns. & Fb. & Uns. & Fb. & Uns. & Fb. & Uns. & Fb. & Uns. & Fb.\\
\midrule
Familiar, no target & 6.5 & 5.3 & 9.6 & 4.3 & 9.5 & 4.4 & 13.5 & 16.0 & 16.8 & 13.0 & 17.9 & 12.7 \\
Familiar, $\alpha=10\%$ & 5.9 & 6.9 & 7.4 & \textbf{6.5} & 8.2 & 8.7 & 9.4 & 22.3 & 9.7 & \textbf{21.4} & 10.3 & 32.4 \\
Familiar, $\alpha=5\%$ & 2.8 & 7.7 & 2.8 & \textbf{7.1} & 3.5 & 31.8 & 4.1 & 36.4 & 4.5 & \textbf{34.9} & 4.5 & 56.6 \\
Familiar, $\alpha=2\%$ & 0.6 & 11.7 & 0.5 & \textbf{10.6} & 0.7 & 60.7 & 1.0 & \textbf{59.9} & 0.9 & 71.5 & 1.2 & 81.8 \\
\addlinespace[2pt]
New combinations, no target & 6.6 & 5.1 & 9.8 & 4.3 & 9.8 & 4.3 & 13.2 & 16.4 & 17.8 & 13.1 & 20.7 & 12.5 \\
New combinations, $\alpha=5\%$ & 2.7 & 7.5 & 2.3 & \textbf{7.1} & 1.8 & 49.2 & 4.3 & \textbf{35.4} & 3.8 & 39.3 & 2.0 & 82.0 \\
\addlinespace[2pt]
Beyond the population, $\alpha=5\%$ & 3.5 & 46.6 & 4.4 & \textbf{44.8} & 3.5 & 60.5 & -- & -- & -- & -- & -- & -- \\
 
\bottomrule
\end{tabular}
\end{table}
\begin{table}[htbp]
\caption{Fixed-penalty policies and outcome interfaces over 20 splits.
Agr.\ is agreement with the oracle without a risk target.
The remaining columns report declined missions at three familiar-mission targets and at 5\% on new combinations (New), all in percent.
Penalty 0 is the plain policy.}
\label{tab:penalty}
\centering\footnotesize
\setlength{\tabcolsep}{3pt}
\begin{tabular}{@{}lrrrrr@{\hspace{10pt}}rrrrr@{}}
\toprule
 & \multicolumn{5}{c}{Unitree Go2} & \multicolumn{5}{c}{Unitree H1}\\
\cmidrule(r){2-6}\cmidrule(l){7-11}
Interface & Agr. & 10\% & 5\% & 2\% & New & Agr. & 10\% & 5\% & 2\% & New\\
\midrule
Penalty 0 & 78 & 8.7 & 31.8 & 60.7 & 49.2 & 67 & 32.4 & 56.6 & 81.8 & 82.0 \\
Penalty 0.3 & 78 & 6.8 & 22.3 & 55.0 & 32.2 & 66 & 27.2 & 52.3 & 80.0 & 74.2 \\
Penalty 1 & 77 & 7.1 & 8.2 & 43.5 & 8.5 & 63 & 24.8 & 44.7 & 77.4 & 56.9 \\
Penalty 3 & 70 & 8.5 & 8.5 & 20.4 & 8.7 & 53 & 40.1 & 40.8 & 72.8 & 46.8 \\
Penalty 10 & 61 & 14.6 & 14.6 & 16.1 & 14.0 & 40 & 58.2 & 58.2 & 71.5 & 64.8 \\
\midrule
Route-outcome model & 78 & 6.5 & 7.1 & 10.6 & 7.1 & 67 & 21.4 & 34.9 & 71.5 & 39.3 \\
Capability contract & 78 & 6.9 & 7.7 & 11.7 & 7.5 & 66 & 22.3 & 36.4 & 59.9 & 35.4 \\
 
\bottomrule
\end{tabular}
\end{table}

\subsection{Damage Beyond the Source Population}
On the 80 more severely damaged Go2 robots, the guarded interfaces select infeasible routes on fewer than 5\% of missions on average, while declining 44.8--60.5\% (Table~\ref{tab:h2h}).
This is an observed result outside the calibration guarantee.
The test examines more severe damage on one platform and does not establish reliable detection of arbitrary distribution shifts.

\subsection{Assessment After Damage}
Table~\ref{tab:ident} compares the trial burden and hazardous-terrain ranking of the probe model with hazard retesting.
AUC is the area under the receiver operating characteristic curve for traversal success, so this comparison tests ranking rather than probability calibration.
The stale model uses an undamaged H1 robot or the average of three near-nominal Go2 robots.
Its AUC is below that of the population average on both platforms.
Hazard retesting uses $T$ trials per terrain at random start states to fit a robot-specific logit offset to the population's start-state profile by maximum a posteriori estimation.

\begin{table}[htbp]
\caption{Assessment of 100 test robots per platform.
Runs and falls are per robot, and AUC measures ranking on hazardous terrain.
Two designed probes match four trials per hazard closely in AUC with fewer runs and falls.}
\label{tab:ident}
\centering\footnotesize
\setlength{\tabcolsep}{4.2pt}
\begin{tabular}{@{}lrrr@{\hspace{14pt}}rrr@{}}
\toprule
 & \multicolumn{3}{c}{Unitree Go2} & \multicolumn{3}{c}{Unitree H1}\\
\cmidrule(r){2-4}\cmidrule(l){5-7}
Strategy & Runs & Falls & AUC & Runs & Falls & AUC\\
\midrule
Stale pre-damage model & 0 & 0.00 & 0.60 & 0 & 0.00 & 0.61 \\
Population average & 0 & 0.00 & 0.67 & 0 & 0.00 & 0.75 \\
Two flat probes & 2 & 0.01 & 0.81 & 2 & 0.00 & 0.84 \\
Two designed probes (ours) & 2 & 0.05 & 0.87 & 2 & 0.07 & 0.86 \\
Re-test each hazard $4\times$ & 36 & 17.5 & 0.87 & 32 & 15.9 & 0.85 \\
Re-test each hazard $8\times$ & 72 & 36.1 & 0.89 & 64 & 31.4 & 0.87 \\
 
\bottomrule
\end{tabular}
\end{table}

Designed probes yield little additional AUC beyond two, whereas random benign probes require about eight to reach comparable scores (Table~\ref{tab:probes}).
This provides the empirical basis for the two-probe budget.

\begin{table}[htbp]
\caption{Hazardous-terrain AUC against probe count $K$.
Random benign probes average ten draws.}
\label{tab:probes}
\centering\footnotesize
\setlength{\tabcolsep}{4pt}
\begin{tabular}{@{}lrrrr@{\hspace{12pt}}rrrr@{}}
\toprule
 & \multicolumn{4}{c}{Unitree Go2} & \multicolumn{4}{c}{Unitree H1}\\
\cmidrule(r){2-5}\cmidrule(l){6-9}
Probes & $K=1$ & 2 & 4 & 8 & $K=1$ & 2 & 4 & 8\\
\midrule
Designed probes & 0.837 & 0.867 & 0.862 & 0.865 & 0.851 & 0.859 & 0.862 & 0.863 \\
Random benign probes & 0.816 & 0.844 & 0.860 & 0.865 & 0.831 & 0.843 & 0.857 & 0.863 \\
Flat-ground probes & 0.768 & 0.811 & 0.823 & 0.828 & 0.839 & 0.842 & 0.839 & 0.847 \\
 
\bottomrule
\end{tabular}
\end{table}

\section{Benchmark Details}
\label{app:bench}
\paragraph{Damage.}
Each robot has zero to three weakened leg joints with probabilities 0.10, 0.35, 0.35, and 0.20, chosen among 12 Go2 joints or 10 H1 leg joints.
Weakening factors in $[0.25,0.80]$ scale the torque--speed envelope on Go2 and actuator torque limits on H1.
The payload perturbation uses a mass multiplier of $1.0$--$1.3$ with a center-of-mass shift up to 3\,cm, and foot friction is in $[0.4,1.0]$.
The 80 additional Go2 robots exceed these ranges in one factor, using four or five weak joints, actuation factors $0.10$--$0.25$, mass factors $1.30$--$1.45$, or friction $0.25$--$0.40$.

\paragraph{Terrains and probes.}
Go2 terrains include flat ground, ascents of 12, 20, and $24^\circ$, a $26^\circ$ descent, stairs with 8, 12, and 14\,cm steps, a 28\,cm descent, rough ground of $\pm14$\,cm, and boxes of 12, 16, and 20\,cm.
H1 terrains include flat ground, ascents of 9--$22^\circ$, descents of 18 and $29^\circ$, stairs of 26--34\,cm, a 44\,cm descent, rough ground of $\pm26$ and $\pm32$\,cm, and boxes of 38--50\,cm.
Terrains with mean source success in $[0.05,0.8]$ are designated hazardous, giving nine on Go2 and eight on H1.
Probe candidates are flat ground, a $12^\circ$ ascent, and a $26^\circ$ descent on Go2, and flat ground, 9 and $12^\circ$ ascents, and an $18^\circ$ descent on H1, each from four start states.
The benchmark contains 145,920 Go2 and 109,072 H1 rollouts.

\paragraph{Successful traversal and falls.}
Success requires covering 1.5\,m on flat ground or slopes and 2.0--2.2\,m elsewhere within 6--10\,s without falling.
A fall is registered when base clearance above terrain drops below 0.14\,m on Go2 or 0.55\,m on H1, when roll or pitch exceeds 55--$60^\circ$ on Go2 or $45^\circ$ on H1, or when the base or torso touches the ground.

\section{Robot and Terrain Examples}
\label{app:examples}
Figures~\ref{fig:go2_examples} and~\ref{fig:h1_examples} show recorded single-terrain runs from the simulation setting.
Each image is a frame from a run, generally at half the required distance or earlier termination.
The annotations identify the robot's damage and the represented terrain.
These snapshots illustrate the physical setting; the route-selection example in Fig.~\ref{fig:concept} uses explicitly illustrative predictions rather than claiming a measured route execution.

\begin{figure}[!htbp]
\centering
\begin{minipage}{0.48\textwidth}
\centering
\includegraphics[width=\linewidth,trim=0.000bp 241.920bp 589.824bp 2.880bp,clip]{figures/go2_terrain_examples.pdf}
{\small Go2 test robot\par}
\end{minipage}\hfill
\begin{minipage}{0.48\textwidth}
\centering
\includegraphics[width=\linewidth,trim=147.456bp 241.920bp 442.368bp 2.880bp,clip]{figures/go2_terrain_examples.pdf}
{\small Probe candidate, $12^\circ$ ascent\par}
\end{minipage}\par\medskip
\begin{minipage}{0.48\textwidth}
\centering
\includegraphics[width=\linewidth,trim=294.912bp 125.280bp 294.912bp 119.520bp,clip]{figures/go2_terrain_examples.pdf}
{\small Hazard, 12\,cm stairs\par}
\end{minipage}\hfill
\begin{minipage}{0.48\textwidth}
\centering
\includegraphics[width=\linewidth,trim=147.456bp 8.640bp 442.368bp 236.160bp,clip]{figures/go2_terrain_examples.pdf}
{\small Hazard, $\pm14$\,cm rough ground\par}
\end{minipage}
\caption{Go2 robot 300 with weakened front left thigh, rear left thigh, and front left calf joints at 0.49, 0.72, and 0.66 times nominal torque, respectively.
Its base mass is 1.25 times nominal and its foot friction is 0.59.
The selected frames show the robot and three terrains from the recorded examples.}
\label{fig:go2_examples}
\end{figure}

\begin{figure}[!htbp]
\centering
\begin{minipage}{0.48\textwidth}
\centering
\includegraphics[width=\linewidth,trim=0.000bp 241.920bp 589.824bp 2.880bp,clip]{figures/h1_terrain_examples.pdf}
{\small H1 test robot\par}
\end{minipage}\hfill
\begin{minipage}{0.48\textwidth}
\centering
\includegraphics[width=\linewidth,trim=442.368bp 241.920bp 147.456bp 2.880bp,clip]{figures/h1_terrain_examples.pdf}
{\small Probe candidate, $18^\circ$ descent\par}
\end{minipage}\par\medskip
\begin{minipage}{0.48\textwidth}
\centering
\includegraphics[width=\linewidth,trim=294.912bp 125.280bp 294.912bp 119.520bp,clip]{figures/h1_terrain_examples.pdf}
{\small Hazard, 30\,cm stairs\par}
\end{minipage}\hfill
\begin{minipage}{0.48\textwidth}
\centering
\includegraphics[width=\linewidth,trim=147.456bp 8.640bp 442.368bp 236.160bp,clip]{figures/h1_terrain_examples.pdf}
{\small Hazard, $\pm26$\,cm rough ground\par}
\end{minipage}
\caption{H1 robot 2100 with weakened left hip pitch, right knee, and right ankle joints at 0.61, 0.28, and 0.42 times nominal torque, respectively.
Its torso mass is 1.05 times nominal and its foot friction is 0.57.
The selected frames show the robot and three terrains from the recorded examples.}
\label{fig:h1_examples}
\end{figure}

\end{document}